\documentclass{article}

\usepackage{graphicx}%
\usepackage{multirow}%
\usepackage{amsmath,amssymb,amsfonts}%
\usepackage{amsthm}%
\usepackage{mathrsfs}%
\usepackage[title]{appendix}%
\usepackage{xcolor}%
\usepackage{textcomp}%
\usepackage{manyfoot}%
\usepackage{booktabs}%
\usepackage{algorithm}%
\usepackage{algorithmicx}%
\usepackage{algpseudocode}%
\usepackage{listings}%

\usepackage{hyperref}
\usepackage[utf8]{inputenc}

\definecolor{bluelink}{RGB}{0,0,238}

\lstdefinelanguage{Rips}{%
  keywords={consts, vars, rules,levels,int,string,bool,float},
  keywordstyle=\color{teal}\bfseries,
  ndkeywords={soft, true, false},
  ndkeywordstyle=\color{bluelink}\bfseries,
  identifierstyle=\color{black},
  sensitive=rue,
  comment=[l]{\#},
  commentstyle=\color{gray}\ttfamily,
  stringstyle=\color{red}\ttfamily,
  morestring=[b]"
}
\lstdefinestyle{rips}{ 
    basicstyle=\scriptsize,
    keywordstyle=\color{red},
    numbers=left,
    numbersep=10pt,
    showstringspaces=false,
    stringstyle=\color{blue},
    tabsize=2,
    language=Rips
}
\theoremstyle{thmstyleone}%
\theoremstyle{thmstyletwo}%

\theoremstyle{thmstylethree}%

\title{Implementing a Robot Intrusion Prevention System (RIPS) for ROS 2}

\author{Enrique Soriano-Salvador \\
\and
Francisco Martín-Rico \\
\and
Gorka Guardiola Múzquiz \\
~\\
EIF, Universidad Rey Juan Carlos
~\\
{\small \texttt{\{enrique.soriano,francisco.martin,gorka.guardiola\}@urjc.es}}
}

\begin{document}
\maketitle

\abstract{It is imperative to develop an intrusion prevention system (IPS), specifically designed
for autonomous robotic systems. This is due to the unique
nature of these cyber-physical systems (CPS), which are not merely
typical distributed systems. These systems employ their own systems software
(i.e. robotic middleware and frameworks) and execute distinct components to
facilitate interaction with various sensors and actuators, and other robotic
components (e.g. cognitive subsystems). Furthermore, as
cyber-physical systems, they engage in interactions with humans and their
physical environment, as exemplified by social robots. These interactions
can potentially lead to serious consequences, including physical
damage. In response to this need, we have designed and implemented RIPS,
an intrusion prevention system tailored for robotic applications based
on ROS 2, the framework that has established itself as the de facto
standard for developing robotic applications.  This manuscript
provides a comprehensive exposition of the issue,
the security aspects of ROS 2 applications, and the key points of the
threat model we created for our robotic environment.
It also describes the architecture and the implementation of
our initial research prototype and a language
specifically designed for defining detection and
prevention rules for diverse, real-world robotic scenarios.
Moreover, the manuscript provides a comprehensive evaluation of the approach,
that includes a set of experiments with a real social robot
executing a well known testbed used in international robotic
competitions.}


\maketitle

\section{Introduction}

Social robots need to be autonomous, thereby facilitating their
deployment in increasingly unstructured environments. They
present vast opportunities for augmenting human economic productivity
and overall well-being. However, they also introduce significant risks
that are challenging for humans to evaluate and manage.  Social robots are
becoming an increasingly common presence in our society, with heterogeneous
applications (e.g. vacuum cleaners, caregivers, security guards, etc.).
The authors have spent time exploring the implications and
interplay between cybersecurity, safety, and explainability in this kind
of systems.  While safety mechanisms have been a long-standing feature in
robotics, security mechanisms have, regrettably, been largely overlooked.
A main conclusion drawn from our previous studies
is the necessity for the development
of specialized intrusion detection systems (IPS) specifically designed for
robotics.

Like others~\cite{guerrero2017empirical},
we assert that traditional
network intrusion detection/prevention systems
are ill-suited for autonomous robotic systems, cognitive
social robots, and other forms of cyber-physical systems (CPS). These
CPS are complex amalgamations of digital, analog, physical, and human
components, all intricately engineered to function through a seamless
integration of physics and logic. In such contexts, while conventional
IPS solutions may be effective in detecting and preventing low-level
communication attacks, they fall short in thoroughly inspecting robotic
communications and implementing safety-related mitigation measures at the
robotic level.

Consequently, we have engineered a novel system, named
RIPS (Robotic Intrusion Prevention System), with the primary objective of
blending safety and security measures. The main ideas behind our system
were defined in a previous work~\cite{rips-journal}.  RIPS
is specifically designed for robotic applications based on
ROS 2 (the current de facto standard for deploying
robotic applications)~\cite{ros2,ros2dds} and
aims to enhance the security measures in place for these advanced robotic
systems. RIPS uses \textit{System Modes}~\cite{9474424},
a safety mechanism for executing operational modes in robotic systems
that enables the segregation of contingency management and the functional setup,
irrespective of the inherent logic of the robotic application.  This
mechanism facilitates alterations in the state of the robot and modifications
in its behavior as a component of the mitigations activated by the RIPS
rules.
For instance, as a consequence of the activation of a specific rule
within RIPS,
the robot could be prohibited from accessing a particular
location and forced to switch off dangerous actuators (e.g. the robotic arm).
This exemplifies the capacity of the system to dynamically adapt its operational
parameters
in response to specific rule-based triggers. This adaptability is a key
feature of the design of the system, enhancing its robustness and flexibility
in diverse operational contexts.

In this manuscript, we introduce the first functional research
prototype of RIPS,
which relies on two main components:
(i) A monitor used to capture the interactions of ROS 2 nodes
following the Publisher/Subscriber model and
(ii) an engine that defines and evaluates
the rules to trigger different mitigation actions.
These rules are specified by a custom DSL (Domain Specific Language)
specifically created for RIPS.
The ROS 2 monitor is implemented in Python,
while the rule engine is developed in Go
and follows an unorthodox approach: It is an interpreter and a transpiler
(i.e. it is able to generate Go code from the specific rules language,
that can be compiled to generate
a native binary executable) for \textit{Rips programs} that define the rules.
The rule language is, in fact, a statically and strongly
typed programming language.

Through the evaluation of our prototype,
we aim to answer the following research questions:
\emph{Is it viable to detect and mitigate cyberattacks in a ROS 2 robotic
application operating on a physical robot? Can it be done by implementing a scheme that
evaluates rules analyzing the structure of the system
and the content of transmitted messages,
and employs operational modes to respond to potential threats?}
The hypotheses are:
\begin{itemize}
        \item H0: The viability of the proposed approach is significantly
        influenced by the number of messages and participants in a real
        robotic application. Specifically, the volume of messages and
        the number of entities involved in the communication process
        are expected to have a substantial impact on the effectiveness
        of the detection and mitigation scheme.

        \item H1: Analyzing the structural composition of the application
        enables the detection of cyberattacks and anomalous behaviors.

        \item H2: Analyzing the type and content of messages within a
        ROS 2 application facilitates the detection of cyberattacks and
        anomalous behaviors.

        \item H3: Upon detecting anomalous behaviors and cyberattacks,
        it is possible to execute appropriate response actions within
        a social robotic application.

        \item H4: Upon detecting anomalous behaviors and cyberattacks,
        it is feasible to respond within a reasonable timeframe to
        mitigate their logical and physical consequences.
\end{itemize}

In summary, the principal contributions of this work are:

\begin{itemize}
	\item A complete exposition of the issue at hand, along
	with an analysis of the security aspects of ROS 2 applications,
	the key points of the threat model we created for our robotic
	environment, and a complete review of the state of the art.

	\item The delineation of an architecture for the Robotic Intrusion
	Prevention System (RIPS) and the description of an
	implementation, centered on the two primary components
	of this system, namely, a ROS 2 monitor and the rule engine.
	These two components collectively constitute
	the core of the RIPS, underpinning its functionality and
	performance.

	\item The description of a custom scheme to define threat levels,
	rules for detection, and mitigation actions,
	including the definition of a language specifically designed for this system
	that provides a flexible and powerful tool for diverse robotic
	scenarios.

        \item A comprehensive evaluation within a real-world scenario
        specifically tailored for social robotics applications and tested
        on an actual robotic platform.
\end{itemize}

The rest of the manuscript is organized as follows:
Section \ref{background} provides the background information on the ROS 2 framework;
Section \ref{related} discusses the related work;
Section \ref{model} summarizes the threat model that we confront, as well
as the specific robotic platform chosen for this project;
Section \ref{implementation} describes  the architecture of our research prototype
and implementation details of the two main components of the system;
Section \ref{ruleslang} describes the language designed for rule definition
and presents some example configurations;
Section \ref{eval} presents the evaluation of the prototype;
and
Section \ref{conclusions} provides the conclusions.

\section{Background \label{background}}

This subsection presents the necessary background information
to understand the general problem.
At present, applications in the field of robotics are characterized as
distributed systems built upon a robotic framework or middleware.
ROS (Robot Operating System) is the most important one.
There are two versions of ROS:
The original ROS~\cite{Quigley2009,quigley2015}
framework
(from now on, we will use the term ROS 1 to avoid confusion)
and ROS 2~\cite{ros2}.
Both facilitate the construction of distributed robotic systems, and are
widely used. However, they are fundamentally different systems and should
not be conflated. The main difference is that ROS 1 is centralized (i.e.
it depends on a master node), while ROS 2 is not.
ROS 2 is gradually supplanting ROS 1, particularly in
research projects. We will focus on ROS 2.

\subsection{ROS 2}

In the ROS 2 framework, a robotic
application is conceptualized as an ensemble of components (i.e. nodes)
that operate across one or more autonomous machines.
Despite the fact that ROS 2 has been ported to a range of operating
systems, the components constituting a robotic system predominantly
execute on Linux systems (as standard processes or be
isolated within containers).
These systems are typically interconnected via a conventional LAN.
This arrangement facilitates the design,
development and execution of the robotic application.

ROS 2 constitutes a collection of libraries and tools
that enable the robotic components to function in unison.
Those components include sensors (cameras, laser, LiDAR, etc.), actuators,
the cognitive subsystem of the robot (e.g. goal dispatcher, planner, etc.),
dialog and speech subsystems, logging subsystems, explainability
components, object recognition components, and so on.

The most prevalent
among these mechanisms is the messaging model, which operates on the principles
of the Publisher/Subscriber paradigm\footnote{For the first prototype of RIPS
we decided to focus on the Publisher/Subscriber
model, even though ROS 2 provides other types of communication
mechanisms, such as \textit{services} and \textit{actions}.
Upon validation of the proposed approach, the rest of mechanisms may be
addressed without much trouble.}. This model facilitates effective
communication and interaction among the nodes.
The programmer defines a set of
\textit{topics} to exchange typed messages (e.g. in frames from a video camera,
commands for a gripper actuator, etc.).
Source nodes send messages to others by publishing in the corresponding topic.
Sink nodes receive the messages by subscribing to the corresponding topic.
The group of nodes, topics, etc. that are active constitutes
the \textit{computation graph}.

ROS 2 is based on the Data Distribution
Service (DDS) middleware~\cite{ros2dds},
specifically on the RTPS (Real Time Publish Subscribe)
protocol specification~\cite{rtps-doc}.
This middleware provides the transport, serialization,
and discovery mechanisms.
ROS 2 provides a
client layer library (the
\texttt{rclcpp} library for C++ and the \texttt{rclpy} library for Python
are the standard implementations) and
an abstract DDS layer (\texttt{rmw}) to interact with the underlying
DDS middleware implementation, that can be provided by different
vendors (see, for example, \texttt{Cyclone DDS} and eProsima's \texttt{FastDDS}~\cite{fastdds-doc}).
Note that different DDS implementations may depend on different
specific services.

The preferred protocol for communication in ROS 2 is UDP
operating in unicast mode\footnote{This is related to the
QoS options used by many projects.}. However, it is noteworthy
that multicast mode can also be employed as an alternative and is better for
performance reasons. Indeed,
the utilization of multicast is paramount for
our system, as will be elaborated upon in Section \ref{model}.

\subsection{Security in ROS 2}

ROS 1 did not include built-in security mechanisms.
It has been object of multiple security analysis (see for example ~\cite{mcclean2013preliminary,portugal2017role,jeong2017study,Dieber2020,yu2021data,rivera2019rosploit,abeykoon2017forensic,demarinis2019scanning,breiling2017,toris2014}.
On the other hand, DDS (and therefore, ROS 2) provides
built-in security mechanisms for:

\begin{itemize}
	\item \textbf{Node authentication.}
	It is based on PKI (Public Key Infrastructure).
	Each node has a key pair and a
	X.509 certificate signed by a Certification
	Authority (CA).
	The authentication protocol uses
	the ECDH algorithm and ECDSA signatures.

	\item \textbf{Data confidentiality.}
	Messages are encrypted with the AES-GCM
	algorithm. A per-topic symmetric key is shared by all the participants
	(publishers and subscribers). There are three modes for data encryption
	(full data encryption, encryption of sub-messages from an entity, and
	encryption of just the payload of the published messages).

	\item \textbf{Message authentication.}
	Messages are authenticated with the GMAC (Galois Message Authentication Code)
	algorithm. Only publishers are able to authenticate messages.
	Each subscriber shares a key with the publisher. All messages published
	in the topic includes the different GMACs computed for all subscribers.

	\item \textbf{Authorization and access control.}
	Nodes include signed
	XML configuration files that include access control rules
	for subscribing to and publishing in the corresponding
	ROS 2 topics.
\end{itemize}

ROS 2 includes some additional tools for configuring those security
mechanisms~\cite{8594462}.
In general, these security capabilities have an acceptable impact
in latency and loss terms~\cite{MARTIN201895}. Nevertheless,
the security mechanisms are disabled by default. When enabled,
they are not enforced by default. In order to enforce the mechanisms, they
must be configured in strict mode (in this mode, if the configuration files
for security are missing or incorrect, the ROS 2 node cannot be executed).
We will provide more details in the next section.

Despite the availability of those mechanisms, ROS 2 developers
often neglect the security aspects.

\section{Related Work \label{related}}

The literature on the security and safety aspects of robotics is vast.
We shall overlook the studies that focus on safety concerns in robotics.
In this section, we will attempt to summarize the bibliography centered
on security for robotics systems, categorizing it into three distinct groups.

\subsection{Security Aspects for Autonomous Robotic Systems}

There are several surveys on robotic security that provide a comprehensive starting point for exploring the security aspects of autonomous robots. Archibald et al. ~\cite{archibald2017survey} categorized the vulnerabilities of robotic systems into several classes: physical, sensor, communication, software, system-level, and user vulnerabilities. They also outlined potential physical and logical attacks. Jahan et al. conducted a survey on the security modeling of autonomous systems~\cite{jahan2019security}. Their work reviewed the historical evolution, approaches, and trends, and analyzed various types of autonomous systems. Yaacoub et al. published a survey on the primary security vulnerabilities, threats, risks, and their impacts, specifically for robotics environments~\cite{archibald2017survey}.

Numerous research works have been conducted on this particular topic. For example,
Lera et al.~\cite{lera2017cybersecurity} published a comprehensive study focusing on cybersecurity attacks related to service robots. The study includes a risk taxonomy specifically designed for these types of robotic applications. Notably, it distinguishes between security threats and safety threats, providing a nuanced understanding of the risks associated with service robotics.
Cerrudo et al.~\cite{cerrudo2017hacking} conducted a comprehensive study on a variety of generic security issues in robotics. These encompass insecure communications, authentication and authorization issues, weak cryptography, privacy concerns, and more. Additionally, they provided illustrative attack scenarios across diverse fields.
Clark et al.~\cite{clark2017cybersecurity} delineated both real and potential threats to robotics at various levels, including hardware, firmware, operating system, and application. The study also includes an analysis of the impact of such attacks. Furthermore, a set of countermeasures for each level was presented, underscoring the importance of comprehensive security measures in robotics. This work contributes significantly to our understanding of the security landscape in robotics.
Basan et al.~\cite{basan2019analysis} proposed a methodology aimed at analyzing the security of a network of mobile robots. This involved classifying the structural and functional characteristics of robotic systems. Additionally, they conducted an analysis of potential attackers, including their objectives and capabilities.
Dutta et al.~\cite{dutta2021cybersecurity} introduced a study centered on intelligent connectivity in IoT-aided robotics. This research delves into the security issues and outlines a methodology for designing control specifications, dividing the security mechanisms into two categories (data-centered and system-centered).

A substantial body of research has been dedicated to exploring the security facets of autonomous vehicles, which is pertinent to the challenge we are currently confronting~\cite{cavthreatssurvey,kim2021cybersecurity,limbasiya2022systematic,kukkala2022roadmap,burzio2018cybersecurity,7866869,stottelaar2015practical,coffed2014threat,hu2009study,9730540,cao2019adversarial,petit2014potential,shin2017illusion,yan2016can,chauhan2014platform,tippenhauer2011requirements,shepard2012evaluation,psiaki2014gnss,petit2015remote,wang2021can,cao2021invisible,hallyburton2022security,brown2017adversarial,lu2017no,eykholt2018robust,kelarestaghi2019intelligent,choi2020cyber}.

\subsection{Security of Robotic Middleware and Frameworks}

In 2008, Mohamed et al.~\cite{mohamed2008} published a survey on middleware
for robotics. They stated, \emph{``the security mechanisms within the middleware
solutions for robotics are inadequately investigated.''}
In those days, few works focused on the security of robotic communications.

\subsubsection{ROS 1 and other middleware}

After 2008, researchers started to focus more on this issues.
In 2013, McClean et al. presented a research focused on
applying low-cost attacks to ROS 1 systems~\cite{mcclean2013preliminary}.
The study is based on the experience with a ROS 1 honeypot system
evaluated in the DEF CON 20 hacking conference.
Morante et al.~\cite{morante2015cryptobotics}
analyzed the security of ROS 1 and YARP. They concluded that
both architectures were not prepared to protect against attacks and raised
questions about the performance of robotic software after enabling
security mechanisms.
Portugal et al.~\cite{portugal2017role} also analyzed the security issues of ROS 1.
They proposed multiple countermeasures for different levels (hardware, network,
operating system, etc.).
Jeong et al.~\cite{jeong2017study} also analyzed and tested several
vulnerabilities of ROS 1. They described node impersonation and replay attacks,
eavesdropping, and service hijacking.
Dieber et al.~\cite{Dieber2020}  studied different
attacks for ROS 1, and developed a penetration-testing tool named ROSPenTo.

Yu et al.~\cite{yu2021data} also studied the vulnerabilities of
ROS 1 and designed threat models for the TCPROS protocol and the
ROS 1 Master/Slave API.
ROSploit~\cite{rivera2019rosploit} is another offensive tool
for footprinting and exploitation of ROS 1 applications.
This tool is modeled after other popular security tools, such
as NMAP~\cite{lyon2008nmap} and Metasploit~\cite{holik2014effective}.
Abeykoon et al.~\cite{abeykoon2017forensic} presented a forensic methodology
for ROS 1 systems, which provides gathering,
identifying and preserving evidences of attacks or incidents.
DeMarinis et al.~\cite{demarinis2019scanning} published the
results of a scanning of ROS 1 nodes in the entire IPv4 address space of
the Internet in 2017 and 2018. They found more than 300 robots
exposed. They also performed some consented attacks as a proof of concept.

Some research efforts were made to add security mechanisms to ROS 1~\cite{breiling2017}.
Toris et al.~\cite{toris2014} proposed the use of web tokens to provide authentication.
Dieber et al.~\cite{dieber2016} designed a security architecture intended for use on top of ROS 1,
based on a dedicated authorization server used to validate participants.
SROS~\cite{white2016,sros} is a set of security enhancements for ROS 1. It provides
transport encryption (based on PKI and TLS) and access control mechanisms.
Secure ROS~\cite{secureros} (not to be confused with SROS) is a fork of the core
packages of ROS 1 with IPSec~\cite{doraswamy2003ipsec}.
It adds encryption to ROS 1 without requiring any alterations to the core API.

\subsubsection{ROS 2 and DDS}

Martín et al.~\cite{MARTIN201895} analyzed the security mechanisms of different
robotic middleware. They presented a quantitative analysis
of the impact of the security capabilities for two popular middleware
used in robotics (Ice and the Fast-RTPS DDS implementation),  and a
comparative of these middleware (with and without security).
Kim et al.~\cite{Jongkilros2} also conducted a study of the ROS 2 security
mechanisms and their performance.
DiLuoffo et al.~\cite{diluoffo2018robot} published a review of ROS 2 security
and potential risks. They also analyzed the impact of the security mechanisms
in performance. They concluded that the degradation is evident from a
configuration with no security to a fully secured configuration, and discussed the
trade-off between performance and security in robotic scenarios.
Fernandez et al.~\cite{fernandez2020performance} analyzed the performance
of the security and QoS (Quality of Service) mechanisms of ROS 2.
Their experiments showed
a measurable latency and throughput change between 50 different scenarios.
The results showed that security mechanisms have a larger influence on performance than
QoS mechanisms.

White et al.~\cite{white2019network} published an attacker model for DDS
that makes use of passive network reconnaissance techniques in
conjunction with formal verification and model checking to
analyze the underlying topology of information flow. They explored
attacks such as DoS, partitioning of the data bus, or
vulnerability excavation of vendor implementations.
Yen et al.~\cite{ddsblackhat21} presented a dissection of the DDS layers and
some vulnerabilities found in some DDS implementations
(OCI OpenDDS, RTI ConnextDDS, Cyclone DDS, and GurumDDS) through fuzzing techniques.

\subsection{Monitors and Intrusion Detection in Robotic Systems}

Several methods have been proposed to detect attacks in robotic environments.
Urbina et al.
published a survey on attack detection methods in industrial control
systems (ICSs)~\cite{10.1145/2976749.2978388}.
Vuong et al.~\cite{7363359} presented a method to detect attacks
by using the data collected from on-board systems and processes,
employing a method based on decision trees.
Guerrero-Higueras et al.~\cite{guerrero2017empirical} carried out an empirical analysis, which revealed statistically significant discrepancies in the data provided by beacon-based Real Time Location Systems (RTLSs) when the robot is subjected to an attack.
Sabaliauskaite et al.~\cite{7423162} worked on statistical techniques
to detect stealthy attacks designed
to avoid detection by using knowledge of the system’s
model.

Next, we will focus on ROS 1 and ROS 2.
To the best of our knowledge, there are no previous intrusion
prevention systems created for ROS 2 that
include a monitor, a tailored language to define rules, and
different mitigation alternatives, including
robotic-specific safety mechanisms like System Modes.
We believe that RIPS stands as a unique solution in this space.

\subsubsection{ROS 1}

ARNI~\cite{bihlmaier2014increasing,bihlmaier2016advanced}
finds configuration errors and
bottlenecks in the communications of ROS 1. It allows the user to establish a reference state for the entire ROS 1 system. It then persistently contrasts this reference state with the actual state. In the event of discrepancies, the user is alerted via a dashboard. Additionally, the user has the ability to specify countermeasures, which the system will subsequently execute.

ROSRV~\cite{rosrv} is a runtime verification framework for ROS 1 applications.
It provides a speciﬁcation language to define safety properties
and access control policies, and enforces them at runtime.

ROSMonitoring~\cite{rosmonitoring} is another framework to support
Runtime Verification (RV)~\cite{LEUCKER2009293} of ROS 1 applications.
It provides a ROS monitor node to intercept all the topics related to the
properties to be verified.

Drums~\cite{monajjemi2015drums} permits the user to monitor the computation graph
of ROS 1 to audit services and communication channels.
Furthermore, it enables the user to oversee host resources, with the capability to store the gathered data in a database for future reference. It also incorporates a data-driven anomaly detection software, Skyline~\cite{skyline}, to detect discrepancies between recent
and historical data.

ROS-defender~\cite{rivera2019ros} integrates a SIEM,
an anomaly detection system (ROSWatch), a intrusion prevention system (ROSDN),
and a firewall for ROS 1.
It is based on Software Defined Networking (SDN).
ROSWatch, conducts inspections of the network and its logs with the aim of detecting intrusions. It employs a pattern matching model to identify attacks in data flows and logs. The rules utilized in this process incorporate identifiers such as nodes and topics, and these can be classified into different priority classes.
ROSDN uses SDN to replace the standard ROS 1 network communication
to filter messages and ban compromised ROS 1.
The main issues of the approach is the dependency on SDN technology
and the poor performance~\cite{rosFM}.

\subsubsection{ROS 2 and DDS}

When it comes to ROS 2, there are fewer works compared to ROS 1:

ROS-FM~\cite{rosFM} is a high-performance monitoring tool for
ROS 1 and ROS 2. It leverages Berkely Packet Filters (eBPF)~\cite{ebpf} and eXpress Data Path (XDP)~\cite{hoiland2018express} as its foundational technologies. The framework also incorporates a tool for enforcing security policies and an application for data visualization,  but it does not
include detection or mitigation mechanisms.

HAROS~\cite{9474545} is a framework specifically designed to debug ROS 2 applications.
It provides a static analyzer that examines the source code and generates a formal model using a
custom language. The  properties extracted from this model are then used for testing purposes.

Vulcanexus~\cite{vulcanexus} is a robust monitoring tool for ROS 2. It encompasses a variety of libraries, tools, simulators, and other tools for ROS 2. It features a dedicated ROS 2 monitor that tracks the performance of communications with the eProsima Fast DDS implementation. The monitor provides real-time data on various aspects of the communications, including packet loss, latency, and throughput. Nevertheless, it does not provide tools for threat detection and mitigation.

\section{Threat Model \label{model}}

This section aims to briefly describe our threat model.
Evidently, the primary asset under consideration is the robot itself
and the physical environment in which it operates (including
objects and living beings).

\begin{figure}[t!]
\begin{center}
\includegraphics[width=9cm]{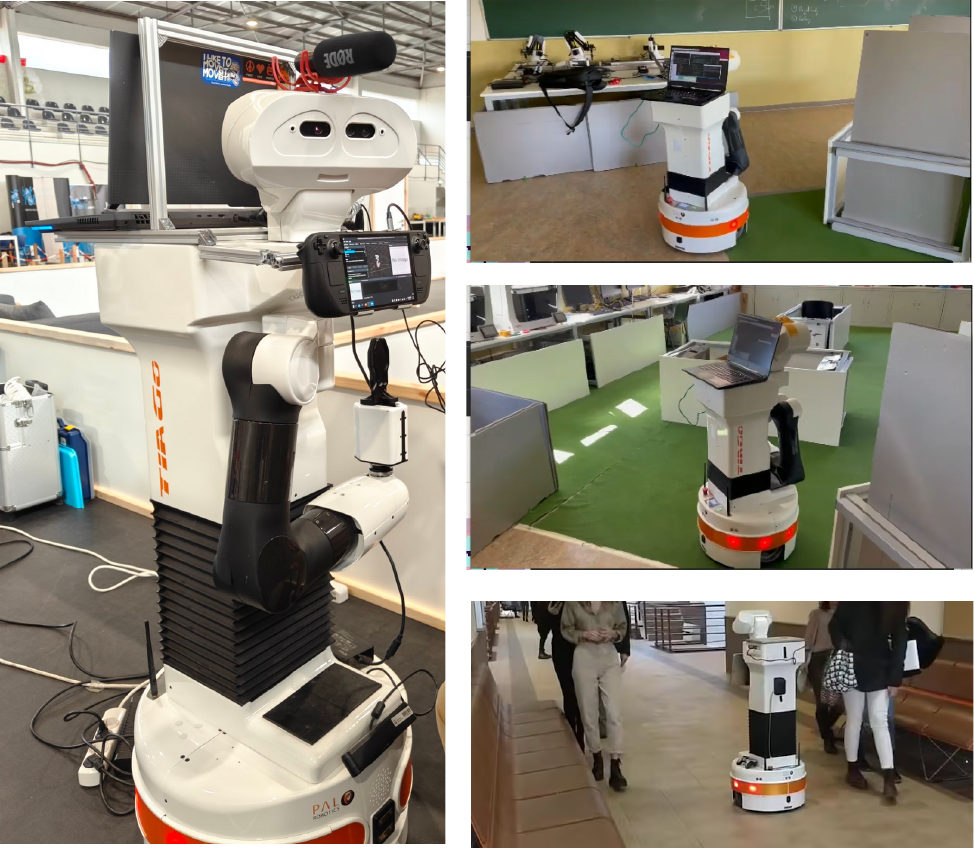}
\caption{Our robot, a TIAGo made by Pal Robotics, operating in
our laboratory and among our students in a public space. \label{fig:tiago}}
\end{center}
\end{figure}

We aim to present a general threat model suitable for current social autonomous
robots. We will particularise it by focusing on our platform, which is
based on a TIAGo\footnote{https://pal-robotics.com/es/robots/tiago}
robot with diverse sensors and actuators (laser, sonars,
Inertial Measurement Unit sensors, RGB-D camera, microphone,
robotic arm, wrist sensor, etc.). This robot, which is depicted
in Figure~\ref{fig:tiago},
has an expansion panel to connect additional devices (e.g. USB cameras, etc.).
All of its robotic resources and the rest of components (that run
in laptops or mini-PCs mounted on the robot) are exported
as ROS 2 components. The machines communicate through an Ethernet LAN,
which is connected to a Wi-Fi network.

We consider that, in robotics,
an autonomous robot operating in public spaces and collaborative scenarios
must provide the service while providing \emph{cyber-physical
security}~\cite{lerarobothreats,4577833},
by ensuring:

\begin{itemize}
	\item The authenticity of the robot's behavior (i.e., the actions of the
	robot are defined by its genuine software components).

	\item The physical integrity of living beings (humans and animals) and objects
	sharing the collaborative environment.

	\item The explainability data of the autonomous system (decisions and actions).

	\item The authenticity, confidentiality, and integrity of the data
	transmitted and stored by the robotic system, guaranteeing the users' privacy.
\end{itemize}

Let’s narrow our focus to the aspects directly related to RIPS,
considering the wealth of existing studies on the risks and threats
in robotic environments (see for example~\cite{lerarobothreats,cerrudo2017hacking,clark2017cybersecurity,basan2019analysis,yaacoub2021robotics}).

\subsection{Threats}

Simplifying, we assume that the adversaries (actors
like external attackers, malicious software, etc.) can:

\begin{enumerate}
	\item  Access the network used by the ROS 2 system to
	spy the communications and remove, duplicate or modify the
	published messages.

	\item  Execute malicious code (or take complete control of) in some
	legit components of the robotic application (i.e.
	ROS 2 nodes), by exploiting vulnerabilities in the operating system
	or the userspace software. Evidently, the RIPS node is excluded.

	\item Alter the physical space in order
	to cause problems while the robot is
	operating (collisions, crashes, falls, etc.) and
	modify the physical environment of the robot
	to inject incorrect sensor data (e.g. laser or LiDAR).
	This point includes
	stealth and covert attacks (i.e., attacks that cause physical
	effects that can not be easily noticed or identified by
	a human observer~\cite{7866869}) such as
	jamming~\cite{stottelaar2015practical,coffed2014threat,hu2009study},
	spoofing~\cite{9730540,cao2019adversarial,petit2014potential,shin2017illusion,yan2016can,chauhan2014platform,tippenhauer2011requirements,shepard2012evaluation,psiaki2014gnss},
	camera blinding~\cite{petit2015remote,yan2016can},
	altering the objects used by the navigation
	software to guide the robot in the space (e.g., signals, images, beacons, cameras, etc.),
	or inject small perturbations on the camera data in
	order to deceive the AI algorithms used by the cognitive
	subsystem~\cite{brown2017adversarial,lu2017no,eykholt2018robust,kelarestaghi2019intelligent}.
\end{enumerate}

\subsection{Assumptions}

RIPS stipulates two prerequisites for the ROS 2 system configuration:

\begin{itemize}
 	\item The ROS 2 security mechanisms for authentication, confidentiality,
	integrity, and access control must be correctly configured and enabled
	in \textit{strict mode}\footnote{The environment variable
	\texttt{ROS\_SECURITY\_ENABLE} must be set to \texttt{true}
	and the variable \texttt{ROS\_SECURITY\_STRATEGY}
	must be set to \texttt{Enforce}.}.
	The nodes must use correct \emph{keystores}: Each node has its own private key,
	and the public keys and certificates needed to communicate with the others.
	The private key of the root certificate (CA) used to sign the rest of certificates
	must not be stored in any component of the system. Messages must be
	fully encrypted.

	\textit{Rationale:} These security mechanisms enable countermeasures
	for attacks related to Threat 1, like spoofing (authenticity), message
	tampering (integrity), information
	disclosure (confidentiality), illegal access (access control), etc.
	RIPS rules rely on the identity of the nodes.
	Thus, unauthorized nodes must not be part of the robotic system.

	Note that, as stated before, legit nodes can be compromised by attackers
	and become malicious. This is part of the scenario we face.
	The administrator will be able to define RIPS rules
	to detect anomalous and malicious behaviors of the participants
	(e.g. incorrect values
	in messages, unexpected participants, incorrect publishers, etc.)
	and trigger prevention actions and mitigations.
	This is the main goal of RIPS.

	\item We assume that the attacker is not able to remove or modify
	messages just for some selected participants: If a message is altered
	for one recipient, it is altered for the rest.
	The robot's network must be an Ethernet and the
	transport layer used by the DDS middleware must
	take advantage of the multicast capabilities\footnote{For example, for the \texttt{FastDDS} vendor, the multicast
	capabilities for UDP can be configured in the profile
	XML file of the nodes (by defining
	\textit{multicast locators}~\cite{fastdds-doc}).}.


	\textit{Rationale:}  If the robotic system
	relies on unicast communications to send the published messages on a topic
	to the corresponding
	subscribers, malicious publishers can send some messages to
	selected nodes, ignoring the rest of subscribers (e.g. the RIPS monitor).
	This trivial attack would bypass the RIPS detection in the robotic system.
	This prerequisite of making all the communications multicast, forces the publisher to
	send only one message through the wire. This message will be necessarily received
	by all the interested nodes (i.e. all the subscribers for the corresponding topic)
	or by none if it is removed.
	This way, if a message published on a topic, it will be available for all the
	participants, including the RIPS monitor\footnote{In addition, taking advantage of the multicast capabilities
	improves the performance
	of RTPS applications with multiple publishers for a topic~\cite{rtps-doc}.}.

\end{itemize}

Executing RIPS without satisfying these two prerequisites
is feasible; however, it is devoid of purpose or meaningful outcome.
Henceforth, we assume these two prerequisites are met.

\subsection{Mitigation}

Our system will be configured by a system administrator
with the aim of safeguarding the specific robotic system.
The final security of the robot environment is, to a significant
extent, contingent upon the proficiency of the administrator.
This particular issue is evidently beyond the purview of this discussion.
The configuration consists on defining rules.
These rules will include expressions to detect deviations in the
behavior of the robotic application. The expressions will be able to inspect:

\begin{itemize}
	\item The \textit{computation graph} (e.g. number of nodes of the system,
	identity of the nodes, number of topics, name of topics, subscribers
	of a topic, etc) in order to detect attacks related to Threat 2.

	\item The metadata of the messages (e.g. inspection of the topic of
	a published message,
	examination of the publishers of the topic, etc.) to detect intrusions
	and illegal behaviors (e.g. a subscriber that becomes a publisher), that is,
	attacks related to Threat 2.

	\item The data (e.g. sensor information, commands, etc.)
	of the messages that are
	published in all topics to detect values out range,
	unexpected or incoherent, malicious payloads, etc.  in order to
	detect attacks related to Threat 3.

	\item External events, like alerts generated by a conventional low
	level IDS/IPS and commands sent by other external applications or the
	administrator, to mitigate attacks related to Threat 1 (which are also
	mitigated by the built-in ROS 2 security mechanisms).
\end{itemize}

In addition, the rules will be able to trigger actions to prevent damages
and mitigate the attack:

\begin{itemize}
	\item Trigger specific safety actions to avoid or minimize damages
	in the physical
	environment, by disabling some capabilities or halt the robot.

	\item Activate low level network mitigations (firewall, etc.)
	through the execution of external commands.

	\item Activate different types of alerts (sounds alarms, text messages,
	spoken messages, etc.) and generate
	logging data and \textit{explainability} information~\cite{explainability-lera}.
\end{itemize}

\section{Implementation \label{implementation}}

\begin{figure}[t!]
\begin{center}
\includegraphics[width=9cm]{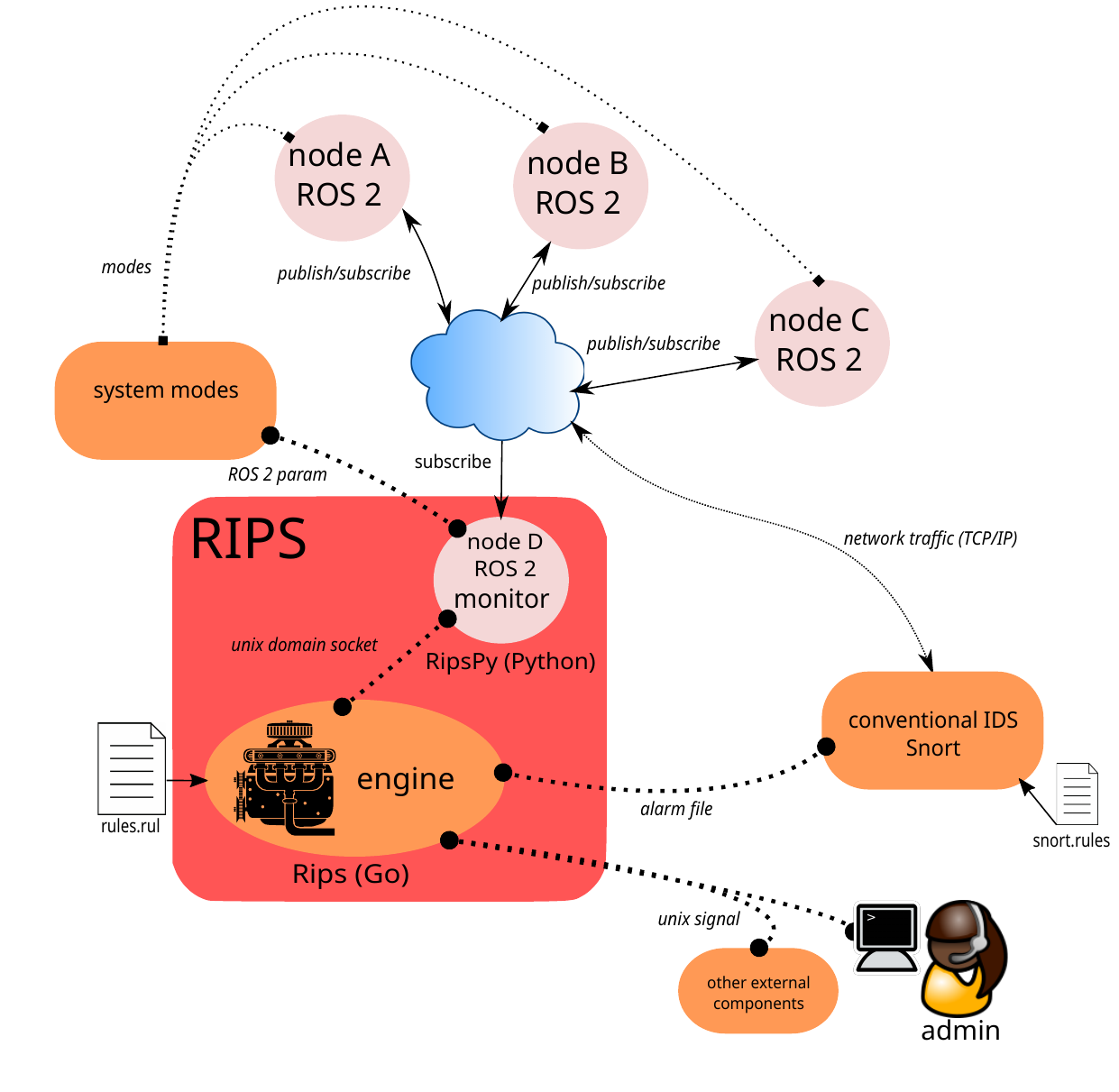}
\caption{The general architecture of our system. \label{fig:arch}}
\end{center}
\end{figure}

The architecture of the system is depicted in Figure~\ref{fig:arch}.
The robotic components, or nodes,
interact by means of the ROS 2 communication mechanisms based on the
Publisher/Subscriber model.
RIPS is a regular ROS 2 participant. It monitors the computation graph
and the published messages, reacting to changes according to the defined rules.

RIPS is formed by two components: the monitor (\texttt{RipsPy})
and the engine (\texttt{Rips}).

\texttt{RipsPy} is a Python 3 program that implements a
ROS 2 node class named \texttt{Ripspy} (derived from the
standard \texttt{Node} class).
As commented in Section~\ref{background}, the standard ROS 2 libraries are
available for Python and C++. We selected Python
due to its capabilities for fast prototyping. This component has around $1000$
lines of code.

\texttt{Rips}, the engine, is an interpreter/transpiler.
It has been implemented in Go~\cite{go},
for several reasons. It is desirable to use a
statically typed language for
a program this complex, so we discarded Python.
The interpreter/transpiler could have been written in C++, but one of the requisites is for
it to be easy to generate a statically linked self contained program
which we can easily cross-compile for different robotic arquitectures
to test the research prototype. In this space, Go excels. We are also
very comfortable writing compilers and interpreters in Go,
as we have written various of different levels of complexity in the past.
The interpreter/transpiler has very few external
dependencies, so it is easy to rewrite
in any language should the need arise later.
This component has around $9000$ lines of code (including tests).

Linking together those two programs would
make things unnecessarily complex for the initial prototype,
so we decided to break the system in two parts
that run in separate processes and communicate through an IPC
mechanism (a Unix domain socket, UDS).

The rules file (\texttt{rules.rul} in Figure~\ref{fig:arch})
defines the rules for RIPS, specified
in the language described in Section~\ref{ruleslang}. A
rule is composed by its name, a boolean
expression and some actions (that are executed when the
expression evaluates to true).

As stated before and explained elsewhere~\cite{rips-journal},
our approach is based on
\textit{System Modes}~\cite{9474424} to change the behavior of the
robot's components in case of intrusion.
RIPS maintains a global \textit{alert level}, that can be
mapped (one to one) to different System Modes
that will define the behavior of the different robotic components.
When the alert level changes, RIPS activates the System
Modes mechanisms to alter the state of the rest of the nodes.

The system can execute external commands to trigger
conventional IPS actions. In addition,
it can receive alarms from a IDS/NIPS/HIPS to detect
low-level network threats (e.g., to detect suspicious messages at
network or transport levels). Our prototype uses Snort~\cite{snort99}
as a low-level IDS/IPS.

External programs can send events to RIPS. This is done through Unix
signals. The engine can catch and recognize such events, so they can
be used by the rules. This may be useful to trigger emergency actions
from a system shell if necessary. The signals \texttt{SIGUSR1}
and \texttt{SIGUSR2} are used for this purpose.

\subsection{Monitor: \texttt{RipsPy} \label{monitor:impl}}

\begin{figure}[t!]
\begin{center}
\includegraphics[width=7cm]{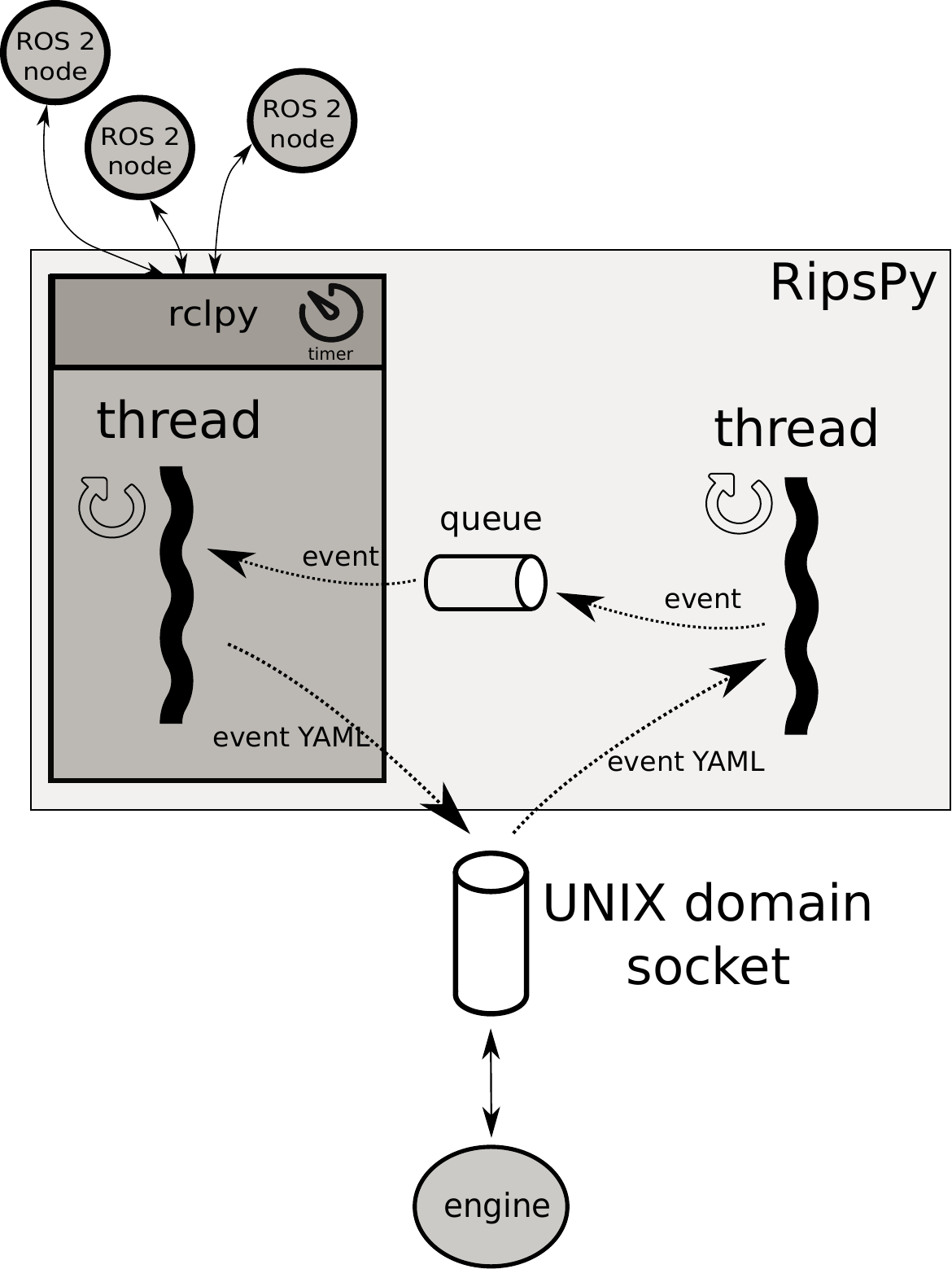}
\caption{The scheme of the monitor, \texttt{RipsPy}. \label{fig:threads}}
\end{center}
\end{figure}

The program creates two threads. Figure \ref{fig:threads} shows the scheme
of the program.
The first thread executes the ROS 2 main loop,
used to receive data from topics and inspect the
computation graph.
Two callbacks (managed by the \texttt{rclpy} library)
are executed by this flow:  (i) the
reception callback to process the messages published in
the different topics and (ii) a timer callback, used to execute
two scheduled tasks:

\begin{itemize}
	\item Updating the current state of the computation
	graph.
	When a new topic is detected, the monitor
	becomes a subscriber if there are rules defined to
	react to the reception of messages. If there is not
	any rule that needs to receive messages from any topic,
	the monitor do not subscribe to any topic in order not
	to impact the performance of the system.

	Moreover, a black list and a white list can be defined.
	The blacklisted topics are ignored by the monitor. For example,
	\texttt{/rosout} (the standard ROS 2 topic used for logging)
	must be always in the black list,
	because it produces feedback loops in the monitor.
	If the white list is not empty, the monitor will only
	subscribe to the specified topics (ignoring the rest).
	The administrator can add new
	topics to these lists, by using two environment
	variables (\texttt{RIPSBLACKLIST}
	and \texttt{RIPSWHITELIST}).

	\item Processing events sent from the engine to
	the monitor. Those events are extracted from
	a thread-safe queue structure.
	There are two types of events: Changes of the current
	RIPS alert level and text alerts (to be displayed in
	a dashboard or the terminal). The events are
	serialized in YAML.
\end{itemize}

The second thread reads messages from the socket
and stores them in the thread-safe queue. This thread does not
use any ROS 2 function. It performs blocking read operations
on the socket.

The two threads only share the thread-safe queue.
The point is to isolate all the ROS 2 activity
(i.e. the interaction with \texttt{rclpy}) in one thread in order
to avoid race conditions.

The monitor sends two different kinds of events to the engine
through the socket, also serialized in YAML:
\texttt{graph} events are sent when the
computation graph changes (i.e. a topic is created
or deleted, a node is created or removed, etc.),
and \texttt{message} events are sent when a message is published in any topic.
The engine reacts to these events, evaluating the corresponding
rules.
Note that the engine is stateless: It does not store the graph.
Events are self-contained (i.e. they must include the computation
graph).

\texttt{RipsPy} periodically executes a method to retrieve the current
computation graph, by using standard ROS 2 methods. The frequency depends
on the polling interval defined by the administrator. This can be configured
with a environment variable named \texttt{RIPSPOLLING}. The default value
is 0.5 seconds. The monitor maintains a
data structure with the discovered topics (with their names, parameters,
subscribers, and publishers) and the nodes (with their names, identifiers,
and services\footnote{Although we focus
in the Publisher/Subscriber communication, the list of
services is also retrieved and saved in the context.}).

For example, the following
\texttt{graph} event describes a context with two
nodes (\texttt{rips} and a node named \texttt{recorder}) and several
topics. It also includes the current alert level and other
relevant information (that will be discussed later):

{\scriptsize
\begin{verbatim}
---
currentlevel: __DEFAULT__
currentgrav: 0.0
lastalert: ''
event: graph
context:
  nodes:
    - node: recorder
      gids:
        - f4.f2.83.53.2d.8d.45.7b.10.20.b5.72.00.00.10.03.00
        - f4.f2.83.53.2d.8d.45.7b.10.20.b5.72.00.00.11.04.00
        - f4.f2.83.53.2d.8d.45.7b.10.20.b5.72.00.00.03.03.00
        - f4.f2.83.53.2d.8d.45.7b.10.20.b5.72.00.00.12.04.00
        - f4.f2.83.53.2d.8d.45.7b.10.20.b5.72.00.00.13.04.00
      services:
        - service: /recorder/describe_parameters
          params:
            - rcl_interfaces/srv/DescribeParameters
        - service: /recorder/get_parameter_types
          params:
            - rcl_interfaces/srv/GetParameterTypes
        - service: /recorder/get_parameters
          params:
            - rcl_interfaces/srv/GetParameters
        - service: /recorder/list_parameters
          params:
            - rcl_interfaces/srv/ListParameters
        - service: /recorder/set_parameters
          params:
            - rcl_interfaces/srv/SetParameters
        - service: /recorder/set_parameters_atomically
          params:
            - rcl_interfaces/srv/SetParametersAtomically
    - node: rips
      gids:
        - 9a.42.c0.75.93.e2.5f.fc.78.54.49.20.00.00.04.03.00.00
        - 9a.42.c0.75.93.e2.5f.fc.78.54.49.20.00.00.11.04.00.00
        - 9a.42.c0.75.93.e2.5f.fc.78.54.49.20.00.00.03.03.00.00
        - 9a.42.c0.75.93.e2.5f.fc.78.54.49.20.00.00.14.04.00.00
        - 9a.42.c0.75.93.e2.5f.fc.78.54.49.20.00.00.15.04.00.00
      services:
        - service: /rips/describe_parameters
          params:
            - rcl_interfaces/srv/DescribeParameters
        - service: /rips/get_parameter_types
          params:
            - rcl_interfaces/srv/GetParameterTypes
        - service: /rips/get_parameters
          params:
            - rcl_interfaces/srv/GetParameters
        - service: /rips/list_parameters
          params:
            - rcl_interfaces/srv/ListParameters
        - service: /rips/set_parameters
          params:
            - rcl_interfaces/srv/SetParameters
        - service: /rips/set_parameters_atomically
          params:
            - rcl_interfaces/srv/SetParametersAtomically
  topics:
    - topic: /parameter_events
      parameters:
        - rcl_interfaces/msg/ParameterEvent
      publishers:
        - recorder
        - rips
      subscribers:
        - recorder
        - rips
    - topic: /rosout
      parameters:
        - rcl_interfaces/msg/Log
      publishers:
        - recorder
        - rips
      subscribers:
        - ~
    - topic: /videocorridor
      parameters:
        - std_msgs/msg/String
      publishers:
        - ~
      subscribers:
        - recorder
        - rips
    - topic: /videooffice
      parameters:
        - std_msgs/msg/String
      publishers:
        - ~
      subscribers:
        - recorder
        - rips
...
\end{verbatim}
}

When the engine notifies a change in
the alert level, the monitor activates the corresponding
\textit{System Mode} state (more information can be found in
previous works~\cite{rips-journal}.)

\subsection{Engine: \texttt{Rips} \label{sect:rips}}

\texttt{Rips} is the program that interprets the rules file.
In fact, the rules file is a program written in our custom language (a
DSL or domain specific language)
and
\texttt{Rips} has two alternative modes of operation: As an interpreter,
it \textit{runs} the program;
As a transpiler, it generates a Go program with the same behavior as
the interpreter running the original program.

In the following example, we will use \texttt{Rips} as an
interpreter (the first path is the socket,
the second path is the scripts directory, that will be discussed later
in Section \ref{sec:levelsstate},
and the last one is the rules file)\footnote{We represent the prompt
of the Linux shell
as \texttt{\$>}}:

~\\

{\scriptsize
\begin{verbatim}
$> rips -s /tmp/sock.777 $RIPSCONFIG/scripts $RIPSCONFIG/rules.rul
\end{verbatim}
}

~\\

Note that, if the scripts directory and the socket are in the default
locations, we could use the \textit{hash bang} invocation (\#!) to execute it
(like any other interpreted program in a Linux system):

~\\

{\scriptsize
\begin{verbatim}
$> chmod +x $RIPSCONFIG/example.rul
$> $RIPSCONFIG/example.rul
\end{verbatim}
}

~\\

In the next example, we will use \texttt{Rips} to transpile the rules file.
The point is to generate a statically linked ELF executable for the engine.
First, we can generate a Go source code file named \texttt{gen.go}:

~\\

{\scriptsize
\begin{verbatim}
$> rips $RIPSCONFIG/example.rul -c $RIPSCONFIG/scripts > gen.go
\end{verbatim}
}

~\\

Then, we can compile it with the Go tools, and execute the \texttt{gen}
program.

In both cases, the \textit{\texttt{Rips} program} will
listen to the Unix domain socket,
read YAML messages from it, and react to events.
The behaviour of both programs other than the performance
is exactly the same.

\section{Rules Language (a.k.a. \textit{\texttt{Rips} programs})\label{ruleslang}}

A \textit{Rips program} is divided
into sections: \texttt{levels}, \texttt{consts}, \texttt{vars}, and \texttt{rules}.
The section marker is a keyword followed by a colon, with the exception of the
sections related to the rules, which have an extra keyword with the type
(\texttt{Graph}, \texttt{Msg}, and \texttt{External}).

Other than the section marker, all the sentences end with a semicolon. Spaces and
new lines are ignored by the parser and only used to separate tokens.
Comments start with \# and end at the end of line.
For example:

\begin{lstlisting}[language=Rips]
levels:
	__DEFAULT__;
	ALERT soft;
	COMPROMISED;
	HALT;

consts:
	MaxNodes int = 5; # rips and 4 participants

vars:
	descalated int = 0;

rules Graph:
	 ! nodecount(1, MaxNodes) && CurrLevel == __DEFAULT__ ?
		alert("detected more than 4 nodes: too many " +
				"nodes, entering level ALERT"),
		trigger(ALERT);

	nodecount(1, MaxNodes) && CurrLevel == ALERT ?
		set(descalated, descalated+1),
		alert("returning to default mode"),
		trigger(__DEFAULT__);

	descalated > 5 && (CurrLevel == ALERT ||
				CurrLevel == __DEFAULT__)?
		alert("too many transitions to alert"),
		exec("/usr/bin/spd-say",
				"too many transitions to alert"),
		trigger(COMPROMISED);

	# should be: rips, monitor & recorder
	! topicsubscribercount("/videocorridor", 0, 3) ?
		alert("videocorridor: too many subscribers"),
		trigger(COMPROMISED);

	# should be: corridorcamera
	! topicpublishercount("/videocorridor", 0, 1) ?
		alert("videocorridor: too many publishers"),
		trigger(COMPROMISED);

	# should be: rips & recorder
	! topicsubscribercount("/videooffice", 0, 2) ?
		alert("videooffice: too many subscribers"),
		trigger(COMPROMISED);

	# should be: officecamera
	! topicpublishercount("/videooffice", 0, 1) ?
		alert("videooffice: too many publishers"),
	trigger(COMPROMISED);

rules Msg:
	topicmatches("/videocorridor") &&
				!publishers("corridorcamera") &&
				CurrLevel != HALT ?
		alert("unauthorized publisher in corridorcamera"),
		exec("/usr/bin/spd-say",
		     "red code the system is totally compromised"),
		exec("/usr/bin/spd-say", "halting the system"),
		exec("/bin/sleep", "5"),
		trigger(HALT);
\end{lstlisting}

This file defines four alert levels (\texttt{\_\_DEFAULT\_\_} is
the initial state), an integer constant is defined, an integer variable,
and some rules (several rules
for graph changes and one rule is for message inspection).
All the rules of this example perform different alert actions and change
the alert level.

\subsection{Levels}

A section marked \texttt{levels} enumerates the possible alert levels.
Each alert level name declared in this section serves
two roles. The name can be passed to a
\texttt{trigger} action to change the current state. As such, it is one of the
possible states and may have extra restrictions (like \texttt{soft},
which limits the legal transitions of the state machine). It also acts
as an integer enumerated value in expressions\footnote{To get a
string with the name see the builtin expression function \texttt{levelname}
described in \ref{sec:predef}.}.  The order of declaration of the levels
defines their severity (i.e.
levels that are declared later are more critical).
We will get back to level transitions later, in Section~\ref{sec:levelsstate}.

An \verb+int+ variable can contain a level (and be used in  \texttt{trigger},  \texttt{levelname}\ldots).
The variable has to be initialized with a level. If the variable gets out of range of the
possible levels, \texttt{trigger} will not change level and \texttt{levelname} will return a zero-length string.

\subsection{Consts and vars}

The available types for literals, variables and named
constants, are: \texttt{string},
\texttt{int}, \texttt{bool} and \texttt{float}.
Internally, integers and floats are kept as 64 bits signed values and
the strings are Unicode.
Literals look like \verb+true+ or \verb+0b11101+.

The section marked \texttt{consts} is used to declare named typed constants.
They have to be initialized with a constant expression. Their lexical scope
starts at their declaration and ends at the end of the file.

The section marked \texttt{vars} is used to declare typed global variables.
They have to be initialized with a constant expression. Their lexical scope starts at
their declaration and ends at the end of the file.  The lifetime of these
variables will coincide with the execution of the program.  \texttt{Rips} will
complain if a variable is unused, unset or never accessed.

\subsection{Rules}

A section marked \texttt{rules} will contain rules to be evaluated when
events are received from the socket: \texttt{Graph} rules react to
\texttt{graph} events and \texttt{Msg} rules react to \texttt{message} events.
In addition, \texttt{External} rules react to external events (signals and alerts
raised by the conventional IDS used by the system).

Each rule contained in a section consists of
a boolean trigger expression, followed by the character '\verb+?+' and a
chain of actions linked by connector operators ('\verb+=>+', '\verb+!>+'
and '\verb+,+'). Alternatively, the Unicode characters
$\rightarrow$ and $\nrightarrow$ can be used
for the arrows\footnote{They are the characters U+2192 and U+219B, which are
encoded in UTF-8 as as 0xe2, 0x86, 0x92  and  0xe2, 0x86, 0x9b respectively.}

The expressions are evaluated upon the arrival of
an event of the corresponding type (i.e. \texttt{Graph}, \texttt{Msg}
or \texttt{External}).
If the expression evaluates to \texttt{true},
then the actions are activated sequentially (depending on the connectors).
Expression functions do not have lateral effects (i.e. they are
similar to mathematical functions).
Expressions can conform the trigger expression of a
rule, and can also be used within the arguments of the actions.

Actions are like procedures (i.e.
they are instructions with lateral effects), although they return a boolean
value: \texttt{true} means success, \texttt{false}
means failure.
Depending on the result of the current action and the connector operator,
the next action is executed:

\begin{itemize}
	\item  '\verb+=>+' means \emph{``execute the
	next action if the result of the previous action was true"}

	\item '\verb+!>+' means \emph{``execute the
       next action if the result of the previous action
       was false''}

       \item '\verb+,+' means \emph{``execute the next action unconditionally''}
\end{itemize}

Of course, in order for the comma connector to evaluate
the following action, the chain has
to arrive to the previous action first without stopping.

Section~\ref{sec:predef} enumerates the builtin expression functions
and actions implemented in the prototype.
Note that some builtins can be used only in
some rule sections (the type system enforces that).

ROS~2 is a complex distributed system and our system is
trying to detect inconsistencies running in a feedback loop with
the whole system. When \texttt{Rips}
detects something suspicious, it communicates
it to the monitor, which may
trigger a system wide reconfiguration because of a change in the
\emph{alert level}. This will feed back new messages with the changes
back to the engine. Too keep all this under control, we must be completely
sure that the program the engine is running is correct. This is
why the compiler is strict, statically and strongly typed, etc.
In general, it tries
to stop on any mistake which may be found statically.

\subsection{Type system\label{sec:types}}

The language is statically and strongly typed.
Variables and constants have to be declared with their types
and their initial value.
Incompatible types cannot be mixed and there is no implicit casting.

The type system of \texttt{Rips} serves two different purposes: preventing
incorrect operation with values
and preventing the builtins to be run in the wrong section.
For example, depending on the type of message received, some builtin
functions can be called but others cannot.

In fact, a type consists of a tuple of two subtypes \verb+(ValueType, ExprType)+.
The first one is the value type (which can be \texttt{string}, \texttt{int}, \texttt{bool}
or \texttt{float}). The other is an expression type, derived from the
kind of event (\texttt{Msg}, etc.). It
is used to control which rule section can use which value.

When an expression is evaluated, the result has
a type compatible only with some types of rule sections.
Thus, it can only be used within them.
On the other hand, actions are universally typed
and can be called from any section.

There are two extra subtypes defined, \texttt{Universal} and
\texttt{Undefined}. Both can be used as value subtypes or as expression
subtypes. They are compatible with any subtype.
The type for two values is compatible if both subtypes in the tuple
(the value type and the expression type)
are compatible, otherwise the operation will result in a type with
one of the subtypes \texttt{Undefined} and the type checking will
ultimately err.
Expressions with simple types like \verb-3 + 4- have a concrete value
type and universal expression type: \verb+(int, Universal)+.

Values of all types are immediate, not references
or pointers. Strings can only be compared lexicographically and are
also treated as values, not pointers. They can also be concatenated with
the $+$ operator,
though they are truncated if they get too long.  The binary and unary
operators that can be used in expressions are roughly the same as in the
C programming language, with the same precedence (with the exception
of shifts which are not present\footnote{
Adding them is probably unwise without adding unsigned types.}).

\subsection{Execution flow}

The program will be interpreted/executed in a loop
(what we call the \textit{main loop} of \texttt{Rips}).
When a event is received through the socket,
a main loop iteration will be executed in this context
in order to evaluate the \texttt{Graph} and \texttt{Msg} rules.

In addition, a main loop
iteration will be executed periodically without any context,
to evaluate the \texttt{External} rules.
In these iterations, \texttt{Rips} checks if a \texttt{SIGUSR1}
or \texttt{SIGUSR2} signal has been received or any low-level IDS
alert has been issued.

\subsection{Alert levels state machine \label{sec:levelsstate}}

The alert levels form a state machine.
The builtin action \texttt{trigger}
changes the current alert level, transitioning to a different state.
By default, the alert level can only rise and de-escalation requires human
intervention.
Nevertheless, if the level is \texttt{soft}, it can be de-escalated.
Levels declared with the \texttt{soft} attribute
can jump to a lower state, but only the one immediately below.

Given this other example (it is a test of \texttt{Rips}):

\begin{lstlisting}[language=Rips]
#!/bin/rips

levels:
   A;
   B;
   C soft;
   D;

consts:
   regexp string ="/pose.*";
   current string = "fff";

vars:
   nmsg int = 0;
   ngraphs int = 0;
   ismatch bool = true;
   lastrule string = "";

rules Msg:
   "zzz" > current ?
     set(nmsg, nmsg + 1), trigger(B);

   topicmatches(regexp) && nmsg > 3 ?
     alert("topic matches") => set(ismatch, true) =>
                                set(lastrule, CurrRule);

rules Graph:
   true ?
     set(ngraphs, ngraphs + 1);

   ngraphs > 10 && ismatch ?
     alert("too many graphs after bad topic rule:" +
     		lastrule), trigger(C);

   ngraphs > 100 && ismatch ?
     alert("alerted too much"), trigger(D);
\end{lstlisting}

\begin{figure}[h]
\begin{center}
	\includegraphics[width=10cm]{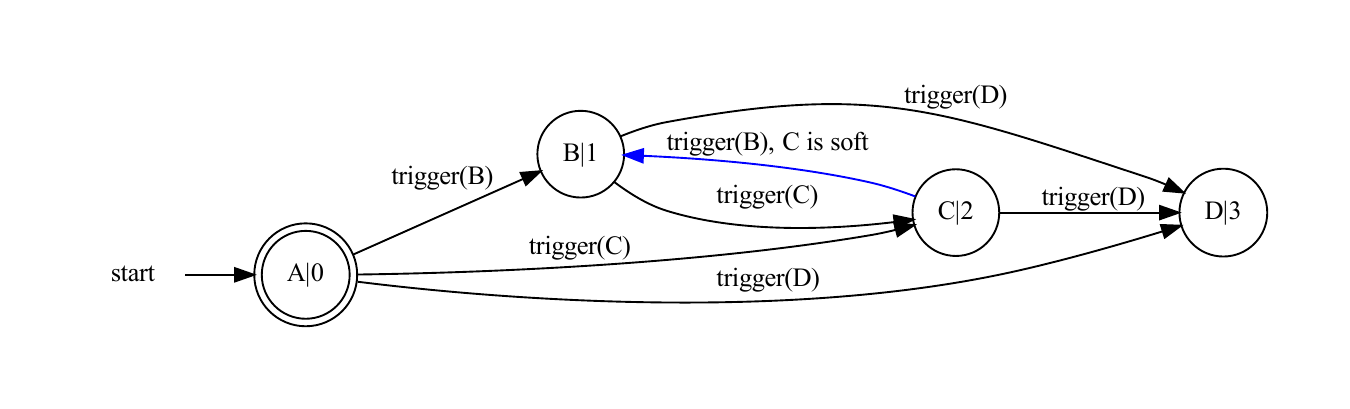}
\caption{State machine for the levels defined in the example file. \label{fig:state}}
\end{center}
\end{figure}

Its state machine
is shown in Figure~\ref{fig:state}.
The initial level is \texttt{A} and that the level \texttt{C} can
transition to \texttt{B} because it has the attribute \texttt{soft}
in the definition (line 6).

Whenever a level is exited and whenever a new level is entered,
an external script is run.
The scripts are named like the level being exited or entered, followed
by the extensions \texttt{.from} and \texttt{.to},
respectively.
The scripts are conventional shell scripts defined by the
administrator\footnote{This scheme is inspired by the traditional Unix \textit{runlevels}.}.

\texttt{Rips} will search for these scripts in a path it receives as
an argument (or by default in \texttt{/etc/rips/scripts}).
It will check that all of the programs are accessible and executable
as part of the compilation process.

In the previous example, for a transition from the level \texttt{C}
to \texttt{D}, a script \textbf{C.from} is executed. Them, the script
\textbf{D.to} is also executed.
When the \texttt{Rips} program starts, only the script \textbf{A.to}
is executed as the initial state is entered.

\subsection{Predefined builtins\label{sec:predef}}

The predefined actions implemented in the prototype
are shown on Table~\ref{table:actions}.

\begin{table}[h!]
	\begin{tabular}{l}
	\\
	 Actions (return bool)\\
	\hline\\
	 set(var, value) \\
	 crash(msg string) \\
	 alert(msg string)   \\
	 exec(path string, args ... string)  \\
	 True(any ... Universal)  \\
	 False(any ... Universal)  \\
	trigger(level int) \\
	\\
	\end{tabular}
	\caption{List of actions.
	Three dots indicate the function is variadic (that is,
	it can accept zero or more arguments in that field). \label{table:actions}}
\end{table}

Their names are self-explanatory.
For example, \texttt{set}
sets a value for a variable.
The builtin action \texttt{crash} receives a string and crashes the interpreter/program
sending the string to everyone (the socket, standard error and wherever the
output is) if possible.
There are two extra builtin actions for debugging the program, \texttt{True} and \texttt{False}.
Both are variadic and receive any type and print it.
They also return \texttt{true} or \texttt{false} respectively, as expected.

The \texttt{exec} action executes an external command in the Linux system.
Consider this \texttt{Graph} rule from the first example of this section:

~\\

{\scriptsize
\begin{verbatim}
descalated > 5 && (CurrLevel == ALERT || CurrLevel == __DEFAULT__)?
        alert("too many transitions to alert"),
        exec("/usr/bin/spd-say", "too many transitions to alert"),
        trigger(COMPROMISED);
\end{verbatim}
}

~\\

If the variable \texttt{descalated} is greater than 5 (it is incremented
by other rules) and the current level is ALERT or \_\_DEFAULT\_\_,
then:

\begin{enumerate}
	\item An alert text message is sent to the monitor.
	\item The Linux command \texttt{spd-say} is executed. This is
	a text-to-speech program. Thus, the message
	``too many transitions to alert'' will be spoken through the speakers
	of the system (the robot!).
	\item The current alert level is changed to \texttt{COMPROMISED}.
\end{enumerate}

\begin{table}[h!]
\begin{tabular}{l}
	\\
	 Msg, about the current message (return bool) \\
	\hline
		msgsubtype(msgtype string, msgsubtype string):\\
		msgtypein(msgtypes  ... string)\\
		payload(pathruleyara string)\\
		plugin(path string)\\
		publishercount(min int, max int)\\
		publishers(pubs  ... string)\\
		publishersinclude(pubs ... string)\\
		subscribercount(min int, max int)	\\
		subscribers(subs  ... string)\\
		subscribersinclude(subs ... string)\\
		topicin(topics ... string)\\
		topicmatches(regex string)\\
		\\
	\end{tabular}
	\caption{List of expressions for \texttt{Msg} rules.
	Three dots indicate the function is variadic.
	\label{table:expressions}}
\end{table}

The predefined expression functions for messages are shown on Table~\ref{table:expressions}.
For example, \texttt{msgtypein} and \texttt{msgsubtype} are used to
check the ROS 2 message type. Other expression builtins accept regular
expressions. For example,
\texttt{topicmatches} inspect the topic of the
received message and returns true if it matches with a regular expression.

The \texttt{plugin} expression is special. It executes an external program,
which can be a C++ or Python program that is able to use the ROS 2 libraries
to examine and interpret the serialized message (which is written to the standard input of the
process that runs the plugin).
The exit status of the Linux process that runs the plugin
determines the return value of the expression (success, an exit status of 0,
means true; and an exit status different than $0$, i.e. failure, means false).
Plugins afford the administrator the opportunity to develop extensions for tailored analysis.

For example, suppose that we have defined this rule in the \texttt{rules Msg} section:

~\\

{\scriptsize
\begin{verbatim}
plugin(pluginpath) && CurrLevel != HALT ?
     alert("plugin detected a bad message"),
     exec("/usr/bin/spd-say", "plugin detected a bad message"),
     exec("/bin/sleep", "5"),
     trigger(HALT);
\end{verbatim}
}

~\\

The expression calls to \texttt{pluggin}, passing the path (a string
constant \texttt{PlugginPath})
of the following pluggin program, which is written in Python:

~\\

{\scriptsize
\begin{lstlisting}[language=python]
#!/usr/bin/python3

import sys
from rosidl_runtime_py.utilities import get_message
from rclpy.serialization import deserialize_message

def main():
  data = sys.stdin.buffer.read()
  msgtype = get_message("std_msg/msg/String")
  try:
    msg = deserialize_message(data, msgtype)
  except Exception as e:
    sys.exit(1)
  else:
    if msg.level == 10 and msg.name == "thename" \
           and msg.msg == "hey" \
           and msg.file == "nofile" \
           and msg.function == "nofunc":
       sys.exit(0)
  sys.exit(1)

if __name__ == '__main__':
    main()
\end{verbatim}
}
 \end{lstlisting}
}

This plugin analyzes the ROS 2 message that have been received. If the type
of the message is \texttt{std\_msg/msg/String} (the standard ROS 2 type for strings).
and its attributes match the requirements
(level is 10, the name is ``thename'', the message is ``hey'', etc.), the expression
function will evaluate to true.

The \texttt{payload} expression is also special.
It employs YARA~\cite{yara} to identify binary patterns
within the payload of the message. The function's parameter is the
path for a YARA rules file.

\begin{table}[h!]
	\begin{tabular}{l}
	\\
	 Graph, about the state of the graph (return bool)\\
	\hline
		nodes(nodes  ... string)\\
		nodesinclude(nodes  ... string)\\
		nodecount(min int, max int)\\
		service(node string, srv string)\\
		servicecount(node string, min int, max int)\\
		services(node string, srvs ... string)\\
		servicesinclude(node string, srvs ... string)\\
		topiccount(min int, max int)\\
		topics(topics ... string)\\
		topicsinclude(topics ... string)\\
		topicpublishercount(topic string, min int, max int)\\
		topicpublishers(topic string, nodes ... string)\\
		topicpublishersinclude(topic string, nodes ... string)\\
		topicsubscribercount(topic string, min int, max int)\\
		topicsubscribers(topic string, nodes ... string)\\
		topicsubscribersinclude(topic string, nodes ... string)\\
		\\
	\end{tabular}
\caption{List of expressions for \texttt{Graph} rules.
Three dots indicate the function is variadic.
\label{table:expressionsgraph}}
\end{table}

The predefined expressions for the graph rules are shown in
Table~\ref{table:expressionsgraph}.
For example, \texttt{topicsinclude} returns true if the current
topics of the graph are included in the set passed as argument,
and \texttt{topicsubscribers} returns true if the subscribers of
a topic are the set specified by the rest of arguments.

\begin{table}[h!]
	\begin{tabular}{l}
	\\
	 External, about external conditions (return bool)\\
	\hline
		idsalert(alert string)\\
		signal(sig string)\\
		\\
	\end{tabular}
	\caption{List of expressions for external events.
	\label{table:external}}
\end{table}

The implemented expressions for external events are shown in
Table~\ref{table:external}.
The expression function \texttt{idsalert} searches for a string
in files whose name match a pattern inside a directory (and its
subdirectories, which are the alert files of the low-level IDS, Snort).
If the string is present, it evaluates to \texttt{true}.

The function \texttt{signal} evaluates to \texttt{true} when a signal
(either ``SIGUSR1'' or ``SIGUSR2'') is received. If
the signal is received multiple times, the expression will evaluate to
\texttt{true} multiple times. There is a counter of signals received and
it will be decremented each time the expression evaluates to \texttt{true},
so signals are not \textit{lost}.

\begin{table}[h!]

	\begin{tabular}{ll}
	\\
	 Universal& return type\\
	\hline
		levelname(level int) & string\\
		string(any Universal) & string\\
		\\
	\end{tabular}
	\caption{List of expressions for external events.
	\label{table:universal}}
\end{table}

Last, the universal expressions implemented by the prototype are
shown in
Table~\ref{table:universal}.
The \texttt{string} builtin returns a string representation of
any type. \texttt{levelname}
receives a level and returns a string with the level name (the
level itself acts as an constant integer).

\subsection{Predefined constants and variables}

Each rule has a constant defined of local lexical scope called
\verb+CurrRule+ with a unique name identifying that rule.

Some variables are predeclared and automatically updated by \texttt{Rips}:

\begin{itemize}
	\item \verb+CurrLevel+ contains the current level, and can be used with \texttt{trigger}  or
	as an integer value in an expression.

	\item \verb+Time+ contains the current time in nanoseconds since the Unix epoch.

	\item \verb+Uptime+ contains the number of nanoseconds which have passed since
	the Rips started.
\end{itemize}

These variables cannot be set (i.e. they are inmutable).

\section{Evaluation \label{eval}}

Assessing a system of this nature presents a significant challenge.
We conducted different kinds of measurements to evaluate it and
respond to our research questions.

We have designed some attacks for an scenario
inspired by a widely recognized application
for social robots used in the \emph{RoboCup@Home} competition~\cite{robocupweb}:
the stage I test called \texttt{receptionist}.
In this test, the TIAGo robot acts as a receptionist for a party:
\emph{``the robot has to take two new guests to the living room to introduce them
and offer a free place to sit; the test takes place in the living room and
the robot starts inside the Arena at a predefined location and
the entrance door is open by default''}~\cite{robocuptestpdf}.
The \emph{RoboCup@Home} competition tests are considered valid testbeds for robotic
applications and systems~\cite{robocuppubs}.

For the following experiments, our TIAGo
robot runs a custom implementation for the \texttt{receptionist} test\footnote{Note that
our team participates in the \emph{RoboCup@Home} competition every year.}.
This  \texttt{receptionist} implementation is based on several ROS 2
components, that execute both in the robot and in an external laptop,
that are connected through an Ethernet network.
Under normal execution, there are more than
56
nodes executing and
213
topics.
A video demonstration for the \texttt{receptionist} without any kind of attack
is available here:

~\\

\begin{center}
	 \href{https://sciencecast.org/casts/sukzxr2ytmv8}{https://sciencecast.org/casts/sukzxr2ytmv8}
\end{center}

~\\

In these experiments,
RIPS is executing in a dedicated laptop, which is also connected to the
Ethernet. The laptop is a
12th generation Intel Core i7-1280P at
4.8 GHz with 48GB of RAM at 3.200MHz running Ubuntu 24.04 and ROS 2
Jazzy. All the experiments have been executed with \texttt{Cyclone DDS}.

First, we discuss the impact of massive monitoring in ROS 2 applications.
Then, we describe 3 experiments performed with the TIAGo robot executing
the \texttt{receptionist} application.
Finally, we present a
quantitative analysis aimed at identifying bottlenecks within the current
prototype. It is important to note that any enhancements to the current
implementation are expected to improve experimental outcomes of RIPS,
thereby establishing our this prototype as a baseline for future
implementations.

\subsection{Monitoring Topics and Global Performance \label{eval:impact}}

In ROS 2 applications, it is highly uncommon for nodes to subscribe to
all topics to receive all the messages published by the rest of the nodes.
Moreover, in some cases, it is absurd to subscribe to all the topics
provided by a node: Some components offer variations of the same resource
through different topics.
For example, in our application, the  node that provides the RGB-D
camera data
offers different topics, to get the camera's feed in different
formats (raw, compressed, etc.).
Under normal operating conditions, it is expected that a node requiring
access to the camera's feed will subscribe to only one of the available topics.
If RIPS were to subscribe to all topics, the resulting load on the camera's
node and the network would be excessive. This consideration extends to other
types of devices and nodes as well.
Furthermore, certain devices are so resource-intensive that they cannot
accommodate multiple subscribers.

Initially, the proposed approach involved subscribing RIPS to all the topics of
the robotic application.
We quickly recognized that this approach might not be feasible for certain
robotic applications.
Monitoring the 213 topics of the \texttt{receptionist} application
significantly affected the performance of the network and the rest of the nodes
(not just \texttt{RipsPy}).
We analyzed the effects of a large number of subscriptions to topics.
The conclusion led to the implementation of configurable white/black list
mechanisms explained in Section~\ref{monitor:impl},
to let the administrator of the system fix
the level of inspection and, therefore, balance the impact in the
performance of the robotic application.

We conducted a series of experiments to determine which topics significantly
influenced the performance of our \texttt{receptionist} application.
We used the \texttt{ros2} tool to measure the
reception rate and bandwidth for a concrete topic\footnote{The command \texttt{ros2 topic hz}.}.
Then, we executed RIPS with \texttt{Msg}
rules and different white list configurations to observe the impact of the
monitored topics in the reception.
With some white list configurations, the reception rate was lower than 2 Hz
and the bandwidth was 1 MiB/s. In other cases, the
rate was greater than 9 Hz and the bandwidth greater than 120 MiB/s.
Through multiple iterations, we determined that limiting subscriptions
to a single topic would result in a degradation of reception
quality: the topic named \texttt{/rgbd\_camera/points} was
too resource-intensive, and it alone can slow down the whole ROS 2
application.
This topic provides a real-time point cloud stream
from a RGB-D camera.
Consequently, within the context of our testbed,
the overhead
associated with monitoring this topic when using RIPS is not acceptable.
In order to conduit the experiments presented in the following
subsections, we blacklisted this topic.

It is important to note that these considerations are specific to the
application and must be thoroughly analyzed prior to securing a ROS 2
system with RIPS.

\subsection{Experiment I: Unauthorized Access to the Front Camera}

In this experiment, we model a scenario wherein an unauthorized
entity gains access to the data stream containing video footage
captured by the robot's front-facing camera. This experiment
aims to replicate a potential security breach where an
illegitimate node subscribes to the topic designated for publishing
the aforementioned video stream, violating the privacy of the party
attendees.

The mitigation action for this experiment is to disable the
front camera (which is not necessary for the robot's navigation) and
alert of the problem.

In the experiment, RIPS detects the intrusion by monitoring the
number of subscribers of the topic that provide the video frames
\footnote{\texttt{/coresense/image\_raw}}.
In our application, the correct number of subscribers for a normal operation
is 4.
The rule file for this detection is:

\begin{lstlisting}[language=Rips]
levels:

    __DEFAULT__;
    COMPROMISED;

consts:

    MaxSubscribers int = 4;
    Topic string = "/coresense/image_raw";

vars:

rules Graph:

    ! topicsubscribercount(Topic, 0, MaxSubscribers) &&
			         CurrLevel == __DEFAULT__ ?
          trigger(COMPROMISED),
          alert("too many subscribers: unauthorized" +
                             " subscriber for the camera"),
          exec("/usr/bin/spd-say", "too many " +
                             "subscribers for the camera"),
          exec("/bin/sleep", "2"),
          exec("/usr/bin/audacious", "-Hq", "/var/tmp/siren.mp3");


\end{lstlisting}

We define two levels for this experiment, \texttt{\_\_DEFAULT\_\_}
and \texttt{COMPROMISED}. There is only one rule: If the number
of subcribers for the corresponding topic is not between 0 and 4,
and the system is not already compromised, the \texttt{COMPROMISED}
level is triggered and some alerts are issued (an alert, a voice message
and a siren sound).
The System Mode for the \texttt{COMPROMISED} level disables
the front camera.
A video demonstration for this experiment is available here:

~\\

\begin{center}
	\href{https://sciencecast.org/casts/1fmhzgidws6q}{https://sciencecast.org/casts/1fmhzgidws6q}
\end{center}

~\\

In this experiment, we measured the number of video frames received by
the malicious node from the moment  the subscription starts to when the camera
is disabled by the System Mode.
We ran the experiments with 2 different values for the polling interval
to retrieve the computation graph from \texttt{RipsPy}: $0.5$ seconds and
$0.1$ seconds. We ran the experiment 10 times for each value. The results
are shown in Table~\ref{table:exp1}.

\begin{table}[h!]
	\begin{tabular}{||c || c | c | c||}
		 \hline
		 Polling Interval (s) & Average (frames) & Median (frames) & $\sigma$ (frames)\\ [0.5ex]
		 \hline\hline
		 0.50 & 24.11 & 25 & 5.46 \\
		 \hline
		 0.10 & 5.66 & 5 & 1.41  \\ [1ex]
		 \hline
	\end{tabular}
	\caption{The average, median and standard deviation
	for the number of video frames received by the malicious node, for
	Experiment I.\label{table:exp1}}
\end{table}

In our robotic platform, the video rate is 30 Hz.
Therefore, when the polling interval is $0.5$ seconds, the malicious node
only retrieves $\approx 0.80$
seconds of video from the front camera.
When
the polling interval is $0.1$ seconds, the malicious node can only retrieve
$\approx 0.18$ seconds of video.

\subsubsection{Experiment II: Unauthorized Navigation}

In the context of this experiment, a restricted zone is delineated
within the living room environment. Under normal operational conditions,
the robot is permitted to enter and navigate within this designated
area. However, in the event that the robot's security is compromised,
it is programmed to be restricted from accessing the restricted zone.
In the event that the robot is located within the restricted zone at
the time of detection, it will be paralyzed.

The attack is characterized as follows: a benign node,
designated as \texttt{recorder}, which is responsible for capturing video
feed from the robot's front camera, transitions into a malicious
state. Subsequently, it attempts to send the robot into the restricted
zone by invoking a ROS 2 action named \texttt{/navigate\_to\_pose}.
Only some authorized nodes of the application are allowed to use this
action (\texttt{goal\_publised}, \texttt{bt\_navigator}, etc.).
The node \texttt{recorder} should not use this ROS 2 action.

In order to detect the attack, RIPS
monitors the topic named:

\begin{center}
  \texttt{/navigate\_to\_pose/\_action/feedback}
\end{center}

Any node that invokes the action \texttt{/navigate\_to\_pose}
is required to subscribe to this topic.
The rule file to detect this attack is:

\begin{lstlisting}[language=Rips]{experiment2.rul}
levels:

     __DEFAULT__;
     COMPROMISED;

consts:

     MonitorizedNode string = "recorder";
     Topic string = "/navigate_to_pose/_action/feedback";

vars:

rules Graph:

      topicsubscribersinclude(Topic, MonitorizedNode)
                            && CurrLevel == __DEFAULT__ ?
          trigger(COMPROMISED),
          alert("unauthorized node for navigation"),
          exec("/usr/bin/spd-say", "unauthorized node " +
                                         "for navigation"),
          exec("/bin/sleep", "2"),
          exec("/usr/bin/audacious", "-Hq",
                                     "/var/tmp/siren.mp3");
\end{lstlisting}

We define two levels for this experiment, \texttt{\_\_DEFAULT\_\_}
and \texttt{COMPROMISED}. Again, there is only one rule: if the
\texttt{recorder} node subscribes to \texttt{feedback}
and the system is not already compromised, the \texttt{COMPROMISED}
level is triggered and some alerts are issued.
The System Mode for the \texttt{COMPROMISED} level forbids the navigation
inside the restricted zone of the living room. In addition, the System
Mode disables the front camera of the robot.

A video demonstration for this experiment is available here:

~\\

\begin{center}
	\href{https://sciencecast.org/casts/x7kva9lpbcf2}{https://sciencecast.org/casts/x7kva9lpbcf2}
\end{center}

~\\

We quantified the elapsed time from the onset of the malicious activity,
specifically the invocation of the action by \texttt{recorder},
to the activation of the System Mode for the
\texttt{COMPROMISED} level. Both events occur in the same machine (the
laptop that executes the \texttt{receptionist} application nodes),
so they use the same clock.
We get the timestamps for both events and calculate the elapsed time.
We also counted the video frames acquired by the malicious node
from the moment the action is invoked
to when the camera is deactivated by the System Mode.
We ran the experiment 10 times.
The polling interval was configured to $0.1$ seconds.
The results are shown in Tables~\ref{table:exp2a} and~\ref{table:exp2b}.

\begin{table}[h!]
	\begin{tabular}{||c || c | c | c||}
		 \hline
		 Polling Interval (s) & Average (s) & Median (s) & $\sigma$ (s)\\ [0.5ex]
		 \hline\hline
		 0.100 & 0.832 & 0.820 & 0.155 \\
		 \hline
	\end{tabular}
	\caption{The average, median and standard deviation
	for the elapsed time from the navigation action execution
	to the activation of the System Mode,
	for Experiment II. \label{table:exp2a}}
\end{table}

\begin{table}[h!]
	\begin{tabular}{||c || c | c | c||}
		 \hline
		 Polling Interval (s) & Average (frames) & Median (frames) & $\sigma$ (frames)\\ [0.5ex]
		 \hline\hline
		 0.10 & 24.88 & 24 & 4.62 \\
		 \hline
	\end{tabular}
	\caption{The average, median and standard deviation
	for the number of video frames received by the malicious node,
	for Experiment II.\label{table:exp2b}}
\end{table}

The results show that the average time to react is $0.832$ seconds.
The TIAGo's maximum speed is $1.5$ m/s~\cite{tiagodatasheet}.
Consequently, the robot is capable of traversing a maximum distance of
1.248 meters before the System Mode intervenes to disable the
navigation function.

As shown in the video demonstration,
the robot was not able to reach the
restricted zone during the experiment (which is delimited with black and
yellow tape within our laboratory setting).

In this case, the malicious node was able to receive $\approx 0.82$
seconds of video before the camera was disabled.

\subsubsection{Experiment III: Malicious Payload}

In this experiment, we defined a new topic named \texttt{/commands}
to publish string messages. The \texttt{recorder} node is receiving
the video from the front camera as in the previous experiment. When it
becomes malicious, it publishes a message in the \texttt{/commands}
topic that includes a malicious payload,
a \emph{shellcode} for Linux AMD64 systems.

To detect the attack, RIPS
monitors all the messages published in the topics of the system.
It uses a \texttt{Msg} rule of type \texttt{payload} to inspect the
data and detect the malicious payload, by using YARA rules.
The rule file for this experiment is:

\begin{lstlisting}[language=Rips]
	levels:
	       __DEFAULT__;
	       HALT;

	consts:

	vars:

	rules Msg:

	      payload("yaraexp3.yar") ?
	            trigger(HALT),
	            alert("malicious payload detected"),
	            exec("/usr/bin/spd-say", "malicious payload detected"),
	            exec("/bin/sleep", "2"),
	            exec("/usr/bin/audacious", "-Hq", "/var/tmp/siren.mp3");
\end{lstlisting}

We define two levels, \texttt{\_\_DEFAULT\_\_} and \texttt{HALT}.
There is only one rule: When any message from any node contains the
malicious payload, the \texttt{HALT} level is triggered.
The \texttt{HALT} System Mode halts the robot.
The YARA rule file to detect the malicious payload is:

\begin{lstlisting}
rule experiment3:
{
   strings:
        # this is the Linux shellcode:
        $payload = "\x31\xc0\x50\x68\x2f\x2f\x73\x68\x68\x2f
                    \x62\x69\x6e\x89\xe3\x50\x53\x89\xe1\xb0
                    \x0b\xcd\x80"

   condition:
        $payload
}
\end{lstlisting}

A video demonstration for this experiment is available here:

~\\

\begin{center}
	\href{https://sciencecast.org/casts/h0aewy479qn6}{https://sciencecast.org/casts/h0aewy479qn6}
\end{center}

~\\

As in the previous experiment,
we measured the elapsed time from the publication of the malicious
message to the activation of the System Mode, and the number of
frames received by the malicious node.
We ran the experiment with 2 different values for the polling interval
($0.1$ and $0.5$ seconds) and two setups:

\begin{itemize}
	\item[$\alpha$)] The whitelist with only one element, \texttt{/commands}.
	In this case, the monitor only subscribes to this topic
	(i.e. the rest of topics are ignored).

	\item[$\beta$)] The blacklist with two topics for the RGB-D
	camera\footnote{\texttt{/head\_from\_camera/depth/points} and
	\texttt{/head\_from\_camera/depth/rgb/points}.}. In this case,
	the monitor subscribes to all the topics except those two
	(which are too load intensive, as explained in Subsection~\ref{eval:impact}).
\end{itemize}

In each case, we ran 10 iterations.
Table~\ref{table:exp3a} shows the time elapsed from the transmission of
the malicious payload to the activation of the System Mode.
Table~\ref{table:exp3b}  shows the number of frames acquired by the attacked
in each case.

\begin{table}[h!]
	\begin{tabular}{||c | c || c | c | c||}
		 \hline
		 Polling Time (s) & Setup & Average (s) & Median (s) & $\sigma$ (s)\\ [0.5ex]
		 \hline\hline
		 0.100 & $\alpha$ & 0.154 & 0.155 & 0.026  \\
		 0.100 & $\beta$ & 3.799 & 1.094 & 23.078 \\
		 0.500 & $\alpha$ & 0.326 & 0.324 & 0.153 \\
		 0.500 & $\beta$ & 1.945 & 1.193 & 2.176 \\
		 \hline
	\end{tabular}
	\caption{The average, median and standard deviation
	for the elapsed time from the navigation action execution
	to the activation of the System Mode, for Experiment III with setups
	$\alpha$ and $\beta$.\label{table:exp3a}}
\end{table}

\begin{table}[h!]
	\begin{tabular}{||c | c || c | c | c||}
		 \hline
		 Polling Time (s) & Setup & Average (frames) & Median (frames) & $\sigma$ (frames)\\ [0.5ex]
		 \hline\hline
		 0.10 & $\alpha$ &  5.00 & 5.00 & 0.86  \\
		 0.10 & $\beta$ & 31.66 & 33.00 & 4.42 \\
		 0.50 & $\alpha$ & 10.22 & 10.00 & 4.84 \\
		 0.50 & $\beta$ & 37.88 & 36.00 & 7.81 \\
		 \hline
	\end{tabular}
	\caption{The average, median and standard deviation
	for the number of video frames received by the malicious node, for
	Experiment III with setups
	$\alpha$ and $\beta$.\label{table:exp3b}}
\end{table}

We can observe that the $\beta$ setup has a considerable impact in
performance: the times are one order of magnitude greater.
Nevertheless, the attacker is not able to retrieve more than $\approx1.23$
seconds of video from the front camera in any scenario.
In the worst case, the maximum distance traversed before halting
would be less than 3 meters.

\subsection{Limitations of the Initial Prototype}

In addition to assessing the system within a real robotic scenario,
our intention was to get certain performance metrics of the prototype,
to evaluate the approach based on two programs
and detect bottlenecks.

To ascertain the relative cost of the different parts of the RIPS engine,
we have built a custom benchmark. The problem evaluating different approaches
is that there can be different sets of rules, different sized messages, etc.
To create a testbed, we captured messages
from a simple scenario, a synthetic system with several nodes, including
RIPS, a publisher which publishes simple values, and a subscriber
which consumes them. We saved this messages into a file, to make the
test reproducible. Then, we ran this scenario against several
\emph{\texttt{Rips} programs}
programs to exercise the different parts of the language.

The concrete scenario and the examples can be found in our public
repository.
We ran the benchmark on a 12th generation Intel Core i7-1280P at
4.8 GHz with 48GB of RAM at 3.200MHz.

\begin{figure}[!t]
\centering
\includegraphics[width=10cm]{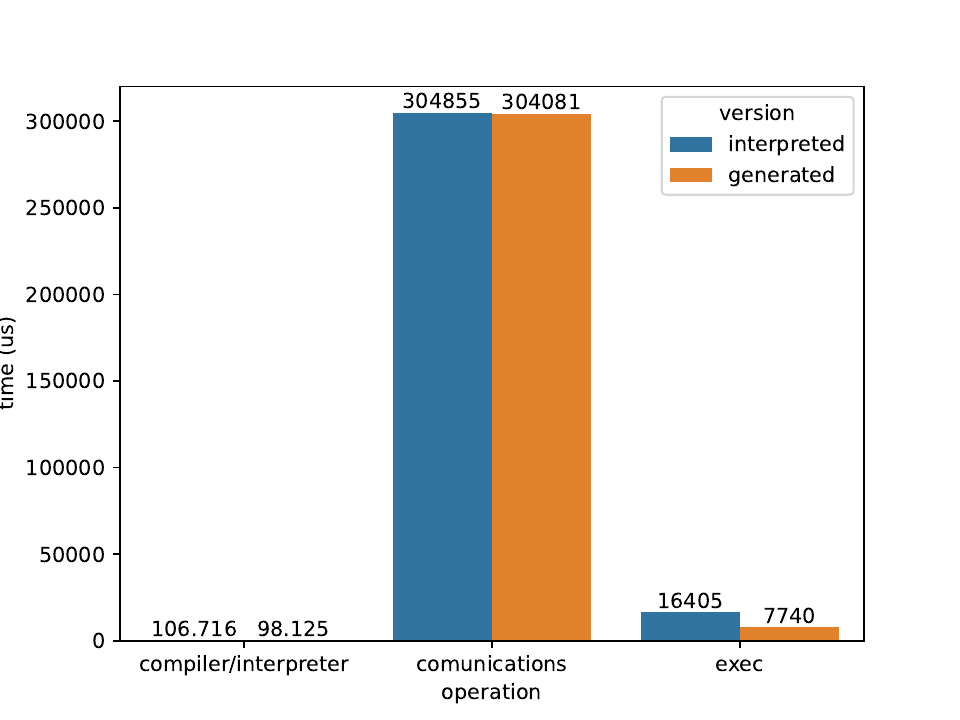}
\caption{Results of the custom benchmark created
to measure the performance of the research prototype.
The numbers show that the communication through the socket and YAML
serialization/deserialization form the bottleneck of the system.}
\label{bench}
\end{figure}

As the graph in Figure~\ref{bench} shows,
the time spent executing the rules (\emph{exec} in the graph) and the
overhead of compilation or execution are dwarfed by the time spent
encoding and decoding the YAML events (\emph{communications operation}
in the graph) sent through the Unix domain socket.
This gives us and idea of the overall cost,
more than the values of the times themselves, which depend on concrete
message sizes and types.
Notice that the times presented by the benchmark may be negligible
when contrasted with the time required for ROS 2 communications (e.g. to
publish a message in a topic) and the execution of actual robotic
actions, like moving the robotic arm.

\subsection{Evaluation Conclusions}

The results explained in~\ref{eval:impact} and Experiment III addressed
hypothesis H0.
Those experimental results demonstrate that, in certain real-world ROS 2
scenarios, it is impractical to monitor all topics (using RIPS or any
other massive subscriber) due to the resulting degradation in network
performance and the performance of the other nodes within the application.
To employ RIPS effectively, it is imperative to adjust the
configuration parameters. This adjustment is contingent upon several
factors, including the specific characteristics of the application,
the quantity of nodes and messages involved, and the underlying
hardware architecture.

The rest of hypotheses were addressed in the experiments with the
\texttt{receptionist} application.
The experiments
indicate that the initial RIPS prototype is fully
operational and the analysis of the structural composition of the
application, and the type and content of the messages, permitted the detection
of the different attacks evaluated.
The performance metrics are sufficient to
handle real-world scenarios for social robots.
In all instances, the mitigation strategies employed are sufficient to
thwart the attack and minimize the associated risks to an acceptable
level and within a reasonable timeframe.

The results of the custom benchmark indicate potential for further optimization.
Note that the cost of serializing and deserializing is just an artifact of
the initial prototype.
Future versions should be one binary executable
that includes both the monitor and the engine, in order to avoid the cost
of serialization and communication.
Another conclusion we extracted from the custom benchmark is that compiling
the rules to native code (\emph{generated} in the graph)
instead of interpreting them, makes the execution much faster.
In this case, it is about two times faster (16405 $\mu$s vs. 7740 $\mu$s).
This evidence underscores the merits of the transpiler-based approach
implemented by \texttt{Rips}.

\section{Conclusions \label{conclusions}}

We have described the first research prototype for RIPS, a
novel Robotic Intrusion Prevention System for ROS 2
robotic systems.
We provided an exhaustive account of the matter in question,
a summary of the threat model that we created for our robotic platform,
and a complete revision of the state of the art.

We described the architecture of the system
and the implementation details of the two main
components of this system, specifically, a ROS 2 monitor implemented in Python
and the rule engine implemented in Go that follows an unconventional approach:
It is an interpreter and a transpiler for a custom domain specific language (DSL)
we have created.
We also described the language, providing some illustrative examples.

Finally, we have detailed a series of experiments conducted within a real
robotic environment, executing a well-known testbed for social robots used
in an international robotics competition (\emph{Robocup@Home}).
Drawing from the empirical data obtained,
we are now positioned to address the research questions posed
in the introduction.
Our first research prototype is operational and in working condition
and there is room for improvement.

Future work includes
the implementation of an unique program that integrates the monitor and
the engine in one binary to solve the bottleneck detected,
the integration with additional safety mechanisms for robotics,
a graphical dashboard application,
and new functions to inspect other ROS 2 mechanisms
(actions and services).

\section*{Availability of Data and Materials}

Data sharing not applicable to this article as no datasets were generated
or analysed during the current study.

RIPS is free software, you can redistribute it and/or modify
it under the terms of the GNU General Public License as published by
the Free Software Foundation. The source code is available in the our
public repositories:

~\\

\begin{center}

 \href{https://github.com/DMARCE-PROJECT/ripspy}{https://github.com/DMARCE-PROJECT/ripspy}

 ~\\

 \href{https://github.com/DMARCE-PROJECT/rips}{https://github.com/DMARCE-PROJECT/rips}

\end{center}

\section*{Funding}

This work is funded under the Proyectos de Generación de
Conocimiento 2021 call of Ministry of Science and Innovation of
Spain co-funded by the European Union, project
PID2021-126592OB-C22 CASCAR/DMARCE.

\section*{Competing Interests}

There are no competing interests in this work.

\section*{Authors' Contributions}

G.G. and E.S. designed and implemented RIPS.
All authors conceived, planned and carried out the experiments
and measurements.
F.M. provided the robotic platform used for the
experiments.
G.G. and E.S. wrote the initial versions of the manuscript.
All authors discussed the results and contributed to the
final manuscript.

\section*{Acknowledgments}

We appreciate the assistance of the staff of the Intelligent Robotics Lab of
the Universidad Rey Juan Carlos, Juan Diego Peña Narváez,
José Miguel Guerrero and Juan Carlos Manzanares,
who collaborated in the experimental setup.

Generative AI software tools (Grammarly\footnote{https://app.grammarly.com} and
Microsoft Copilot\footnote{https://www.bing.com/chat})
have been used exclusively to edit and improve the quality of human-generated
existing text.

\bibliographystyle{IEEEtran}
\bibliography{related}

\end{document}